%% file: m547.tex
\documentclass{ecai} 

\usepackage{latexsym}
\usepackage{amssymb}
\usepackage{amsmath}
\usepackage{amsthm}
\usepackage{booktabs}
\usepackage{enumitem}
\usepackage{graphicx}
\usepackage{color}

\usepackage{array}
\usepackage{multirow}
\usepackage[]{subcaption}
\usepackage{algorithm}
\usepackage{algorithmic}

\newcommand{\BibTeX}{B\kern-.05em{\sc i\kern-.025em b}\kern-.08em\TeX}

\def\BibTeX{{\rm B\kern-.05em{\sc i\kern-.025em b}\kern-.08em
    T\kern-.1667em\lower.7ex\hbox{E}\kern-.125emX}}

 \newcommand\MyBox[3]{
   \fbox{\lower0.75cm
     \vbox to 1.2cm{\vfil
       \hbox to 2cm{\hfil\parbox{1.9cm}{\centering #1\\#2\\#3}\hfil}
       \vfil}%
   }%
}

\definecolor{cerulean}{rgb}{0.0, 0.48, 0.65}

\begin{document}


\begin{frontmatter}


\paperid{547} 


\title{Error-Driven Uncertainty Aware Training}


\author[A,B]{\fnms{Pedro}~\snm{Mendes}\thanks{Corresponding Author. Email: pgmendes@andrew.cmu.edu}}
\author[B]{\fnms{Paolo}~\snm{Romano}}
\author[A]{\fnms{David}~\snm{Garlan}}

\address[A]{Software and Societal Systems Department, Carnegie Mellon University}
\address[B]{INESC-ID and Instituto Superior Tecnico, Universidade de Lisboa}


\begin{abstract}
    \input{sections/1_abstract}

\end{abstract}
\end{frontmatter}

\input{sections/2_intro}

\input{sections/3_RelatedWork}
\input{sections/4_euat}

\input{sections/5_results}

\input{sections/6_conclusion}


\begin{ack}
This work was supported by the Fundação para a Ciência e a Tecnología (Portuguese Foundation for Science and Technology) through the Carnegie Mellon Portugal Program under grant SFRH/BD/151470/2021 via projects with reference  
UIDB/50021/2020 and 
C645008882\mbox{-}00000055.PRR, 
by the NSA grant H98230-23-C-0274,
and by the Advanced Cyberinfrastructure Coordination Ecosystem: Services \& Support (ACCESS) program, where we used the Bridges-2 GPU and Ocean resources at the Pittsburgh Supercomputing Center through allocation CIS220073, which is supported by National Science Foundation grants \#2138259, \#2138286, \#2138307, \#2137603, and \#2138296.
\end{ack}


\bibliography{biblio}

\end{document}

%% file: sections/1_abstract.tex
Neural networks are often overconfident about their predictions, which undermines their reliability and trustworthiness. In this work, we present a novel technique, named Error-Driven Uncertainty Aware Training (EUAT), which aims to enhance the ability of neural classifiers to estimate their uncertainty correctly, namely to be highly uncertain when they output inaccurate predictions and low uncertain when their output is accurate. The EUAT approach operates during the model’s training phase by selectively employing two loss functions depending on whether the training examples are correctly or incorrectly predicted by the model. This allows for pursuing the twofold goal of i) minimizing model uncertainty for correctly predicted inputs and ii) maximizing uncertainty for mispredicted inputs, while preserving the model's misprediction rate. We evaluate EUAT using diverse neural models and datasets in the image recognition domains considering both non-adversarial and adversarial settings. The results show that EUAT outperforms existing approaches for uncertainty estimation (including other uncertainty-aware training techniques, calibration, ensembles, and DEUP) by providing uncertainty estimates that not only have higher quality when evaluated via statistical metrics (e.g., correlation with residuals) but also when employed to build binary classifiers that decide whether the model’s output can be trusted or not and under distributional data shifts.

%% file: sections/2_intro.tex
\section{Introduction}
\label{sec:intro}

Deep Neural Networks (DNNs) have achieved remarkable performance across various domains and are increasingly utilized to automate intricate decision-making processes. 
However, a critical limitation of current neural models is their tendency to display overconfidence in their predictions~\cite{survey_uncertainty,dun}. This overconfidence persists even when erroneous predictions are made, ultimately compromising the reliability and trustworthiness of the models.

Recent research efforts~\cite{MCdropout,CeW,variance_out,mcmc,BayesNN} have been dedicated to enhancing the trustworthiness of DNNs by estimating the model's predictive uncertainty through various approaches.
%
%
Bayesian neural networks (BNNs)~\cite{BayesNN,BNN,BNN1} offer an elegant framework for modeling uncertainty~\cite{bayes_unce}. However, while BNNs provide theoretically sound uncertainty estimates, they incur prohibitive costs, being impractical for large datasets and complex models. To mitigate these challenges, various approximations have been introduced. For instance, Monte Carlo (MC) dropout~\cite{MCdropout}, which leverages dropout regularization during both training and inference stages to approximate the behavior of BNNs. 

Furthermore, numerous studies~\cite{cals,deup,isotonic_regression,ce_pe,ece} have focused on calibrating the models' predicted uncertainty in different ways. These works can be categorized into two primary groups based on whether they: 1)  account for the uncertainty during training by introducing an additional term in the loss function to quantify the model's predictive uncertainty, or ii) implement a post-processing stage to calibrate the model's predicted probabilities using a validation set. 
Although post-processing methods have empirically proven to be effective and cost-efficient~\cite{cals}, they present additional calibration parameters that are sensitive to the method and data used. On the other hand, despite being more expensive, learning-based methods have achieved better performance for uncertainty estimation~\cite{focal_loss,focal_loss1}. 

In this work, we mix both approaches by introducing Error-Driven Uncertainty Aware Training (EUAT), a specialized training procedure for classification tasks that aims at improving the model's uncertainty estimation by imposing high uncertainty for erroneous outputs and low uncertainty for accurate predictions.
To achieve this two-fold goal, during training, EUAT iterates between two loss functions depending on whether the training examples are correctly or incorrectly predicted by the model. 
More in detail, our approach extends a base loss function, which aims to minimize the classification error rate (e.g., cross-entropy (CE)), with an additional term whose objective is to maximize the model's uncertainty for misclassifications and minimize uncertainty for correct classified inputs. 
However, to separate the correctly and incorrectly classified inputs and speed up the training procedure, we first consider pre-train the model and then we apply EUAT to conduct a post-learning-based phase to improve its uncertainty.

We conducted an extensive evaluation of EUAT on classification tasks using popular image recognition models and benchmarks, where we compared our approach against several state-of-the-art methods for uncertainty estimation using six different evaluation metrics. 
Further, we extended our assessment to a binary classification problem, which presents a particularly interesting case involving the class inversion of the high uncertainty outputs that are likely to be wrong classified. 
We also evaluate our technique in an out-of-distribution detection task, where corrupted inputs are used to evaluate the model, and at last, we integrate our function into adversarial training settings in order to identify possible misclassifications based on uncertainty. 
We detail the challenges encountered in each domain/task. 
In general, EUAT presents the best performance in more than 60\% of the metrics considered, and in the majority of the cases where the baselines are more competitive, EUAT is still able to achieve similar performance metrics.
Further, in all the scenarios considered, we show that our strategy can better separate wrong and accurate predictions based on uncertainty, increasing the reliability and trustworthiness of the models.

%% file: sections/3_RelatedWork.tex
\section{Related Work}
\label{sec:related_work}

In this section, we first review different formulations of the problem of estimating models' uncertainty, and the corresponding metrics, proposed in the literature. Subsequently, we analyze existing methods to estimate the uncertainty of DNNs. Finally, we discuss methods aimed at improving uncertainty estimation by adjusting the model's outputs via post-processing or uncertainty-aware training techniques.

\paragraph{Problem definition and metrics.}

The problem of accurately estimating model uncertainty has been formalized using two main theoretical frameworks.
One such formalization is based on the notion of calibration, which aims at aligning the probabilities output by the model with the true likelihood of the predicted outcomes~\cite{model_calibration}. An alternative formalization is based on the  Uncertainty Confusion Matrix (UCM)~\cite{unc_conf_matrix,acc_unc_calib}, as defined in Table~\ref{tab:unc_confusion_matrix}. The UCM specializes the concept of confusion matrix to evaluate the ability to leverage the model's uncertainty to discern correct predictions.
For example, different metrics such as the expected calibration error (ECE)~\cite{ece}, adaptive calibration error~\cite{ace}, or test-based calibration error~\cite{tce}, have been proposed to measure the calibration error of a model. 
On the other hand, leveraging the UCM, several works~\cite{acc_unc_calib,unc_conf_matrix} have exploited additional metrics such as uncertainty accuracy (uA) and the uncertainty area under the curve (uAUC) to enhance the reliability of uncertainty estimates.

\input{sections/5.0_confusionMatrix}

\paragraph{Uncertainty Estimation.} 
One important foundation of these works lies in the computation of uncertainty. Uncertainty in DNNs plays a key role in quantifying the reliability and robustness of their predictions. There are two main types of uncertainty: epistemic uncertainty, associated with the model's lack of knowledge or data, and aleatoric uncertainty, linked to the inherent randomness and unpredictability within the training data~\cite{unc_AT}. Various metrics are employed to measure these types of uncertainty such as predictive entropy (PE) and mutual information (MI)~\cite{pe_mi,pe_mi1}.
However, quantifying uncertainty with DNNs is a challenging task. Bayesian methods~\cite{BayesNN,BayesNN1} can directly provide an estimate of the uncertainty by parameterizing the parameters of the network with distributions. However, training BNNs usually comes with a prohibitive cost.  
Thus, several approximations have been developed. Monte Carlo (MC) dropout~\cite{MCdropout}, which is one of the most popular techniques for uncertainty quantification on DNNs~\cite{uncertainty_survey,mcdropout1}, 
adopts a Bayesian approximation to estimate the uncertainty by sampling multiple dropout masks and aggregating the predictions. Various statistical metrics of the aggregated predictions (such as PE, MI, and variance~\cite{CeW,variance_out,variance_out1}) can then be used to quantify models' uncertainty. 
Additionally, Markov Chain Monte Carlo (MCMC) methods~\cite{mcmc} offer another avenue for drawing the posterior distribution (albeit, those require a long time to converge to the final distribution~\cite{uncertainty_survey,BayesNN}).
Further, Variational Inference (VI) is a technique used to estimate the uncertainty of BNNs by approximating the posterior distribution over the model's weights, which are treated as random variables with associated probability distributions. Training the network to approximate these distributions enables the capture of inherent uncertainty in the model's predictions. 
Moreover, one can also resort to models that can directly output uncertainty estimations (such as Deep Gaussian Processes~\cite{deepGP}), or use Laplace approximations for uncertainty ~\cite{laplace}, or deep ensemble methods that offer yet another approach to estimate model uncertainty~\cite{deep_ensembles,uncertainty_survey}, by aggregating the outputs of each learner in the ensemble and computing relevant metrics such as the entropy or MI.
At last, DEUP~\cite{deup} trains a new simple estimator to predict the uncertainty of the base model exploiting its error/loss, and DUN~\cite{dun} leverages the outputs of different layers of a DNN to measure the uncertainty.

In this work, we resort to MC dropout to estimate the uncertainty of our models and compute the PE using the outputted distributions. 
Distinctly, we explicitly aim to increase the uncertainty of misclassifications by maximizing the PE of the wrong prediction, while minimizing the global error rate.

\paragraph{Post-processing Calibration Methods.} 
Several works investigated how to calibrate a pre-trained model by applying a post-processing phase to optimize model calibration, i.e., to align the output probabilities with the true likelihood of the predicted outcomes of events occurring~\cite{model_calibration}. 
Several techniques, such as Platt Scaling~\cite{platt}, Isotonic Regression~\cite{isotonic_regression}, Temperature Scaling~\cite{model_calibration}, or Beta Calibration~\cite{beta_calib}, can be employed to fine-tune the probabilities outputted by the model ensuring a more accurate alignment with true outcome frequencies.
Krishnan et al.~\cite{acc_unc_calib} introduced an accuracy versus uncertainty calibration (AvUC) loss function devised to obtain well-calibrated uncertainties while simultaneously preserving or enhancing model accuracy, and they extended their approach by proposing a post-hoc calibration phase that combines Temperature Scaling and AvUC.
In addition, Karandikar et al.~\cite{soft_calibration} proposed an extension of ECE and AvUC by developing a soft version of the binning operation underlying these calibration-error estimators, and also extended their approach for post-processing calibration by optimizing the temperature parameter in the temperature scaling method based on the soft calibration error.
Complementary, Gupta et al.~\cite{bin_free_calib} presented a binning-free calibration approach.
However, these calibration methods have been shown to be sensitive to both the model and the validation set~\cite{cals} and perform sub-optimally when faced with shifts in data distribution~\cite{calib_ood}.

\paragraph{Uncertainty Aware Training Methods.}
One fundamental aspect of training DNNs is the choice of a loss function. 
Although Cross Entropy (CE) stands out as a common choice for addressing classification problems, it tends to increase the over-confidence of the resulting model~\cite{model_calibration,focal_loss1}.
Thus, since accounting for predictive uncertainty during training improves model calibration~\cite{acc_unc_calib}, several loss functions, such as focal loss~\cite{focal_loss} or label smoothing~\cite{focal_loss}, have been developed that extend existing ones by incorporating additional terms addressing model uncertainty. 
Similarly, Shamsi et al.~\cite{ce_pe} proposed two loss functions that extend the CE by adding a new term to address the model's uncertainty that can be determined through the PE or the ECE.
Einbinder et al~\cite{Uncertainty_loss1} introduced an uncertainty-aware conformal loss function by adding a new term that quantifies uncertainty via conformal prediction. 
CALS~\cite{cals} exploits the Augmented Lagrangian Multiplier method to adaptively learn the weights of the penalties to balance each term in the new loss function.
Separately, class uncertainty-aware (CUA) loss~\cite{CUA} tailored object detection introduces the uncertainty of each class to augment the loss value when prediction results are uncertain. 
Ding et al.~\cite{Uncertainty_loss} developed an uncertainty-aware loss for selective medical image segmentation that considers uncertainty in the training process to directly maximize the accuracy on the confident segmentation subset, rather than the accuracy on the whole dataset.
Differently from the aforementioned works, our approach takes a distinct path by focusing on leveraging a pre-trained classification model to deliberately increase the uncertainty associated with incorrectly classified inputs without degrading the overall error rate.

Further, our novel method aims to empower Machine Learning (ML) systems to recognize possible model misclassification in production and take customized actions accordingly. This idea can be further extended to adversarial training scenarios, where the deliberate increase in uncertainty for misclassified adversarial inputs enhances the system's ability to detect and respond to potential attacks in production.

%% file: sections/5.0_confusionMatrix.tex
\begin{table}[t]
    \caption{Uncertainty Confusion Matrix}
    \label{tab:unc_confusion_matrix}
    \centering
    \begin{tabular}{c >{\bfseries}l @{\hspace{0.7em}}c @{\hspace{0.4em}}c }
      \multirow{12}{*}{{\bfseries \rotatebox{90}{Correctness}}} & & \multicolumn{2}{c}{\bfseries Uncertainty} \\[1mm]
      & & \bfseries Certain & \bfseries Uncertain  \\
      & \rotatebox{90}{\hspace{-6mm}Correct}  & \MyBox{True}{Certainty}{TC} & \MyBox{False}{Uncertainty}{FU} \\[3.2em]
      &\rotatebox{90}{\hspace{-6mm}Wrong} & \MyBox{False}{Certainty}{FC} & \MyBox{True}{Uncertainty}{TU}
    \end{tabular}
    \vspace{4mm}
    
\end{table}

%% file: sections/4_euat.tex
\section{Error-Driven Uncertainty Aware Training}

\label{sec:euat}

This section introduces a new loss function, which we named Error-Driven Uncertainty Aware Training (EUAT), which aims to enhance the uncertainty estimation capability of a pre-trained model by leveraging a key principle: increasing the uncertainty associated with misclassifications while reducing the uncertainty of correct predictions and error rate of both correctly and incorrectly classified inputs. 

The pseudo-code of EUAT is shown in Algorithm~\ref{alg:algorithm}. We start the process by querying a pre-trained model to determine which inputs of its \textit{training} set are wrongly and correctly classified (see lines~\ref{alg:query1} and ~\ref{alg:query2}). This leads to creating two sets, one containing the incorrect classified inputs $\mathcal{W}$ and the other containing the correct ones $\mathcal{C}$. We use stratified sampling on the target class in order to produce two subsets of equal size. Since we assume to use EUAT on pre-trained models, where typically the error rate on the training set is lower than 50\%, i.e., $|\mathcal{C}|>|\mathcal{W}|$, this procedure typically entails sampling the set of correctly classified inputs ($\mathcal{C}$) in order to reduce its size to match the size of the set of mispredictions ($\mathcal{W}$).  
Then, in order to reduce overfitting, we mix wrong and correct classified inputs and ensure that the batches used for training with EUAT contain a balanced number of samples from $\mathcal{C}$ and $\mathcal{W}$.

EUAT employs distinct loss functions for each set. Since our objective is to deliberately increase the uncertainty of misclassifications, we minimize the CE and maximize the uncertainty  for the wrong-classified inputs while, for the correct-classified inputs, we minimize the CE and the uncertainty, i.e., 
\begin{equation}
    L_{\text{EUAT}}(f_{\theta}(\textbf{x}),\textbf{y})\text{=}
        \begin{cases}
        L_{\text{CE}}(f_{\theta}(\textbf{x}),\textbf{y})\text{-}L_{U}(f_{\theta}(\textbf{x}),\textbf{y}) &\forall \langle \textbf{x},y \rangle \in \mathcal{W}\\
        L_{\text{CE}}(f_{\theta}(\textbf{x}),\textbf{y})\text{+}L_{U}(f_{\theta}(\textbf{x}),\textbf{y}) 
    \hspace{2mm} &\forall \langle \textbf{x},y \rangle  \in \mathcal{C} 
    \end{cases}
    \label{eq:euat}
\end{equation}
where the cross-entropy loss is given by
\begin{equation}
    L_{\text{CE}}(f_{\theta}(\textbf{x}),y) = - \frac{1}{K} \sum_{i=1}^{K} t(x_i) \log (f_{\theta}(x_i)),
    \label{eq:ce}
\end{equation}
($t(x)$ denotes the true label given the input $x$ and $K$ represents the number of dimensions of $\textbf{x}$), and the uncertainty loss is measured by resorting to predictive entropy $H$ 
\begin{equation}
    L_{U}(f_{\theta}(\textbf{x}),y) = H[P(y|\textbf{x)}] = - \sum_{y \in \mathcal{Y}} P(y|\textbf{x}) \log P(y|\textbf{x}),
    \label{eq:pe}
\end{equation}
where $P(y|\textbf{x})$ is the model's output distribution over the set of possible outcomes $\mathcal{Y}$ obtained via MC dropout, namely by approximating the model's output predictions using the average across parameters $\theta_i$  sampled from a dropout distribution  \begin{equation}
    p(y|\mathcal{D},\textbf{x})  \approx \frac{1}{N} \sum_{i=1}^N p(y|\theta_i,\textbf{x}).
\end{equation}  
We resort to MC dropout, since, as mentioned in Section~\ref{sec:related_work}, this approach has been shown to yield a more accurate estimation of the model's uncertainty~\cite{MCdropout}.

EUAT also includes a term that aims to minimize uncertainty for correctly classified inputs, which, intuitively, aims to instill confidence in the model in the regions of the input space where the model is already operating correctly.
Further, by pushing the model's uncertainty for correct and incorrect predictions in opposite directions, EUAT aims to ease the discrimination of erroneous and correct predictions via the uncertainty of the model's forecasts.  
It should also be noted that this proposed loss function is differentiable and continuous, thus enabling its use with classical back-propagation-based training methods.

Note that at the end of each iteration of the while loop  (line~\ref{alg:while}), the model's weights are updated. This affects the model's predictions and, consequently, also the sets $\mathcal{C}$ and $\mathcal{W}$, which is the reason why we update these two sets upon each iteration. 
Additionally, EUAT can be combined with commonly used stopping algorithms to determine when to stop the training process. 
%


\input{sections/algorithm}

%% file: sections/algorithm.tex
\begin{algorithm}[tb]
    \caption{Pseudo-code to train a model with \textsc{EUAT} loss function.}
    \label{alg:algorithm}
    \begin{algorithmic}[1]
    
    \STATE \textbf{Input}: model $f$, training set $\mathcal{S}$, optimizer opt , batch size $B$

    
    \WHILE {\textsc{StopCondition}() $\neq$ True} \label{alg:while}

    \STATE $\mathcal{C}$ = \{$\langle x, y \rangle \in \mathcal{S}$ s.t. $y$=$f(x)$\} \COMMENT{Correct predictions set} \label{alg:query1}
    \STATE $\mathcal{W}$ = \{$\langle x, y \rangle \in \mathcal{S}$ s.t. $y\neq f(x)$\} \COMMENT{Mispredictions set}  \label{alg:query2}
    

    \STATE $\mathcal{C}$ $\leftarrow$ Select at random a subset of $\mathcal{C}$ of size |$\mathcal{W}$| 

    \FOR{$i$ = 1, $\dots$, $\lceil$(|$\mathcal{W}$|+|$\mathcal{C}$|)/$B\rceil$}

    \STATE \COMMENT{$\mathcal{W}_i$ and $\mathcal{C}_i$ have size $B$/2, so the $i$-$t$h batch has size $B$} Let $\mathcal{W}_i$ and $\mathcal{C}_i$ be the $i$-$th$ mini-batch of $\mathcal{W}$ and  $\mathcal{C}$, resp. 


    \STATE \COMMENT{Loss for $\mathcal{W}$:  CE-PE} $L_{\mathcal{W}}(f(\textbf{x}),\textbf{y})$=$L_{CE}(f(\textbf{x}),\textbf{y})$-$L_{U}(f(\textbf{x}),\textbf{y})$  $ \forall (\textbf{x},\textbf{y})$$\in$$\mathcal{W}_i$ 

    \STATE \COMMENT{Loss for $\mathcal{C}$:  CE+PE} $L_{\mathcal{C}}(f(\textbf{x}),\textbf{y})$=$L_{CE}(f(\textbf{x}),\textbf{y})$+$L_{U}(f(\textbf{x}),\textbf{y})$  $ \forall (\textbf{x},\textbf{y})$$\in$$\mathcal{C}_i$ 

    \STATE  $L(f(\textbf{x}),\textbf{y})$=$L_{\mathcal{W}}(f(\textbf{x}),\textbf{y})$+$L_{\mathcal{C}}(f(\textbf{x}),\textbf{y})$ \COMMENT{Add losses}

    \STATE $L$.backward()  \COMMENT{Compute and backpropagate the gradient}
    \STATE opt.step() \COMMENT{Gradient descent updating model's parameters}







     \ENDFOR
     \ENDWHILE
    \STATE  \textbf{return} $f$
    \end{algorithmic}
\end{algorithm}

%% file: sections/5_results.tex
\section{Evaluation}
\label{sec:results}

In this section, we report the evaluation of the EUAT on a variety of domains and tasks.

\subsection{Experimental Setup, Benchmarks, and Baselines}
\label{sec:exp-setup}
\paragraph{Datasets and Models.} In our experimental study, we employed four models and datasets widely used in the image recognition domain namely, ResNet50~\cite{resnet} with ImageNet~\cite{imagenet}, Wide-ResNet-28x10~\cite{wide_resnet} with Cifar100~\cite{cifar10}, ResNet18 with Cifar10, and ResNet18 with SVHN~\cite{svhn}.
We also considered a binary classification model (using ResNet18 with Cifar10 to verify if there is a cat in an image), and an out-of-distribution (OOD) detection task, where corrupted inputs using distributional data shifts are used to evaluate the model. All these models make use of dropout layers and we configure them to use a dropout rate of 0.3. Finally, we evaluate our approach in adversarial training settings using the three models/datasets mentioned above in the image recognition domain.

\paragraph{Baselines.} We compared EUAT against the CE loss, model calibration, DEUP~\cite{deup}, an ensemble of five learners~\cite{deep_ensembles}, CALS~\cite{cals}, and a loss function incorporating both CE and PE (CE+PE)~\cite{ce_pe}. 
To calibrate the model and train DEUP's additional error predictor, we created a validation set comprising 10\% randomly selected samples from the dataset. Further, we resort to Isotonic regression~\cite{isotonic_regression} to calibrate the model, which we have confirmed to achieve superior results when compared to other methods like Platt scaling~\cite{platt}, temperature scaling~\cite{model_calibration}, and beta calibration~\cite{beta_calib}.
Although we experimented DEUP with different validation set sizes, to ensure fairness, we maintained consistency by employing the same validation set size in both cases. For a fair comparison, we ensure each epoch has seen the same amount of data.

\paragraph{Evaluation metrics.} We evaluate the different baselines using six different metrics.
First, we report the uncertainty accuracy (uA) (Eq.~\ref{eq:ua}) and the uncertainty area under the curve (uAUC), which are computed based on the Uncertainty Confusion Matrix~\cite{unc_conf_matrix} defined in Table~\ref{tab:unc_confusion_matrix}.
\begin{equation}
    uA = \frac{TC+TU}{TC+TU+FC+FU}
    \label{eq:ua}
\end{equation}
We also evaluate the models using the correlation between the residuals of the model and predicted uncertainties (Corr. w/ res.)~\cite{deup}, and the Wasserstein distance~\cite{Wasserstein_distance} of the uncertainty distribution between the sets of correct and wrong predictions (Wasser. dist.).
At last, we report the ECE and the model's error. The uncertainty/confidence of the models is always computed via MC dropout using the normalized PE, except for DEUP, which resorts to the loss values of the base model to estimate the quality of its predictions (for a fair comparison, after testing the model trained with DEUP, we had to normalize the loss values).  


\paragraph{Additional implementation details.} To train the models, independently of the considered solution, we use stochastic gradient descent to minimize the loss function using a momentum of 0.9 and a batch size of 64 for all the models, a learning rate of 0.01, and a weight decay of $10^{-5}$ for ResNet50/ImageNet and 0.1 and 0 for the remaining ones, respectively. 

Before training the model using EUAT, we pre-trained the models using CE loss and then decreased the learning rate by $10^3\times$ when applying EUAT. Additionally, we exploited automatic mixed precision to train the ResNet50/ImageNet and  Wide-ResNet/Cifar100. 
The models were trained during 60 epochs (except in the binary classification problem where it was trained for 200 epochs). More in detail, we pre-trained the models for 30 epochs (100 epochs in the binary classification problem) before starting the second phase of training where we applied EUAT (for all the remaining epochs). We use a validation set (corresponding to 10\% of the dataset) in order to tune the uncertainty threshold that is used to discriminate the correctly and incorrectly classified inputs (this threshold is necessary, e.g., to compute the Uncertainty Confusion Matrix). Also, in EUAT, we save the model produced at the end of each training epoch, evaluate its performance on the validation set, and select the one that maximizes the considered performance metric. For fairness, also for the other baselines, we use the same validation-based method to optimize the tuning of the uncertainty threshold as well as the model's selection during the training phase.

The implementation of the training pipeline and additional information to ensure the reproducibility of results are provided in the public repository\footnote{\url{https://github.com/pedrogbmendes/EUAT}}.
All the models and training procedures were implemented in Python3 via the Pytorch framework and trained using a single Nvidia RTX A4000.




\input{sections/5.1_table}

\input{sections/5.1_plots}

\input{sections/5.2_binary}

\input{sections/5.3_ood}

\input{sections/5.4_tableAT}

\subsection{Experimental results}
Next, we report the results obtained using EUAT in the different domains evaluated. 

\subsubsection{Image Recognition Models}
We start by reporting in Table~\ref{tab:results} the results obtained using four models/datasets for image recognition: ResNet50/ImageNet, Wide-ResNet/Cifar100, ResNet18/Cifar10, and ResNet18/SVHN. Across all baselines and metrics,  EUAT consistently demonstrates superior performance, outperforming all the other methods in 16 out of 24 cases.
Notably, in the cases where alternative approaches are more competitive than EUAT, the performance differences are marginal (e.g., the error using EUAT increases by 3.4\%, 13.8\%, and  14.6\% compared to the best baseline training a Wide-ResNet/Cifar100, ResNet18/Cifar10, and ResNet18/SVHN, respectively).

EUAT presents the best uA when training a ResNet50 with ImageNet and a Wide-ResNet with Cifar100, and yields a minimal reduction of 0.5\% and 0.3\% in the uA when training a ResNet18 with Cifar10 and SVHN compared with the best baselines (namely, ensemble and CALS, respectively). 
Additionally, EUAT consistently outperforms existing approaches in terms of uAUC and the correlation between the residuals of the model and predicted uncertainties. EUAT achieves an improvement on the uAUC by up to 15.1\%, 8.9\%, 64.3\%, 28.1\%, 8.5\%, and 20.4\% compared to CE, calibration, DEUP, deep ensemble, CALS, and CE+PE, respectively. 
Moreover, the correlation between the model's residuals and predicted uncertainties improves by 20\%, 20.7\%, 8.6\%, and 11.5\% when using EUAT for training ResNet50 with ImageNet, Wide-ResNet with Cifar100, and ResNet18 with Cifar10 and SVHN compared to the best-performing baseline in each scenario.

Next, the effectiveness of EUAT in distinguishing correct predictions from misclassifications based on the predicted uncertainty is assessed using the Wasserstein distance of the uncertainty between the sets of correct and wrong predictions. More in detail, on average across all models/datasets, the  Wasserstein distance increases by up to 1.7$\times$, 1.3$\times$, 9.7$\times$, 2.3$\times$, 1.4$\times$, and 2.2$\times$ using EUAT compared to CE, calibration, DEUP, ensemble, CALS, and CE+PE, respectively.   
At last, we evaluate the impact of using EUAT on the ECE and models' misclassification rate. EUAT results in lower ECE in two cases (ResNet50/ImageNet and ResNet18/SVHN), while a slight increase is observed in the other two (Wide-ResNet/Cifar100 and ResNet18/Cifar10).
Importantly, the error rate remains consistent across all benchmarks for all baselines, with a notable improvement observed in the ResNet50/ImageNet benchmark, showcasing a reduction of 13.2\% (compared to the best baseline, namely, a deep ensemble).
Furthermore, the large gains of EUAT were obtained using larger models and datasets, where the model is less accurate. In these settings, there exist a relatively larger set of incorrectly classified inputs. This, in turn, allows EUAT to use relatively larger/richer training sets, which, ultimately, increases the effectiveness of the proposed method.




Further, in Figure~\ref{fig:results}, we plot the distribution of the normalized uncertainty of the correctly and incorrectly predicted sets using the different baselines. 
By visualizing these distributions, we verify an improvement in the separation of the uncertainty of these two sets using EUAT (which is confirmed by the computation of the Wasserstein distance in Table~\ref{tab:results}). Overall, these results demonstrate that EUAT   significantly enhances the ability to leverage model's uncertainty to discriminate between correct and incorrect predictions. This ability is a key requirement to build trustworthy AI systems, i.e., to identify in which scenarios the output of an ML model should not be trusted and should, therefore, undergo requiring further scrutiny (e.g., review by human evaluators).



\subsubsection{Binary Classification Problem}

Next, we proceed to assess the effectiveness of the EUAT in a binary classification scenario. In these settings, when the model prediction has high uncertainty, one can take that information as a signal to invert the model's prediction and, thus, output the opposite class. We compared the models obtained using the different baselines and tuned the uncertainty threshold as described in Section~\ref{sec:exp-setup}. Additionally, we conducted an extensive evaluation utilizing supplementary metrics such as F1-score, precision, True Positive Rate (TPR), and True Negative Rate (TNR). Further, we report the error rates both when flipping and not flipping high uncertainty predictions.

Table~\ref{tab:binaryCifar10} presents a comprehensive overview of these performance metrics. Remarkably, the EUAT outperforms other baselines across 9 out of 11 evaluated metrics. More in detail, when considering the geometric mean of the relative improvement of EUAT with respect to all the considered baselines,  EUAT attains 7.4\%, 3.2\%, and 9.1\% enhancements in the F1-score, precision, and TPR, respectively. Interestingly, when flipping highly uncertain predictions, EUAT is the only solution for which the error rate drops significantly when compared to the case of no flipping. This demonstrates that EUAT achieves superior performance with respect to the considered baselines to produce high/low model uncertainty in the presence of wrong/correct predictions. Further, still considering the flipping of high uncertainty prediction, the geometric mean of the relative error rate reduction of EUAT vs. all the considered baselines is 25.6\%.


Furthermore, our method improves the uA, uAUC, and the correlation with the model residuals by up to 9.9\%, 52.5\%, and 113.3\%, while the Wasserstein distance of the uncertainty between the correct and wrong predicted sets is enhanced by up to 17.5$\times$ (with the geometric mean of 2.1$\times$ across all benchmarks). 
Lastly, it is noteworthy that the ECE achieved using EUAT aligns closely with other baselines, namely CE, DEUP, Ensemble, and CALS, and reduces the geometric mean across baselines by 10.7\%, while the misclassification rate lowers by 4,4\%. 
These findings highlight the benefits deriving from the use of EUAT to improve the model's predictive quality in binary classification tasks.



\subsubsection{Out-Of-Distribution Detection Task}

In this section, we assess the ability of EUAT to estimate the model's uncertainty in OOD settings, namely when there are strong distribution shifts with respect to the dataset used to train. These are settings which, in prior works, e.g.,~\cite{soft_calibration}, have been shown to be very challenging for existing uncertainty estimation methods.

More in detail, in this section, we evaluate the effectiveness of EUAT to detect OOD examples based on the predicted uncertainty. 
For each baseline, we trained a ResNet18 using the Cifar10 dataset and then tested it using a corrupted version with Gaussian noise of the same dataset (called Cifar10-C~\cite{cifar10-c}). For a fair comparison, all the compared techniques use a validation set of the same size (10\%) based on the clean inputs of the original test set.  


In Table~\ref{tab:ood}, we compared EUAT against the other baselines considering the aforementioned metrics. Notably, in five out of the six considered metrics, EUAT performs better than the baselines. More in detail, EUAT improves the uA by 9.2\%, 21.9\%, 22.2\%, 0.25\%, 9.9\%, and 10.7\% compared to CE, calibration, DEUP, ensembles, CALS, and CE+PE, respectively, while the uAUC increases by 17.7\%, 43.9\%, 56.7\%, 8.4\%, 11.2\%, and 20.1\%. We also verify an enhancement in the correlation between model residuals and the predicted uncertainty, in the  Wasserstein distance of the uncertainty between the correct and wrong predicted sets, and in the ECE of 2.6$\times$, 4.4$\times$, 2.42$\times$ (geometric mean) compared to the other baselines. 
Finally, all the baselines, except CALS, yielded a model with a larger error rate than EUAT.

\subsubsection{Adversarial Training}

Finally, we evaluate  EUAT in adversarial training settings. 
We opted to exclusively train our models with adversarial examples, utilizing FGSM~\cite{adv_train} to generate perturbations, with a predefined perturbation bound $\epsilon$ set to 4/255.  Due to resource constraints and the overhead introduced by adversarial training, we did not deploy the ResNet50/ImageNet benchmark in adversarial settings.

In Table~\ref{tab:results_at}, we report the results of the different baselines when using as test set either clean or adversarial data.
Overall, even with adversarial data, we observe similar trends compared to the ones already seen in the standard training (Table~\ref{tab:results}). As expected the adversarial error increases in all baselines and benchmarks considered.
In half of the scenarios/metrics assessed, EUAT outperforms the baselines. Quantitatively, employing EUAT yields an increase in the geometric mean of the uAUC (across all baselines) of 16.3\%, 8.3\%, and 10.1\%, when training a Wide-ResNet on Cifar100, a ResNet18 on Cifar10, and a ResNet18 on SVHN datasets, respectively. Moreover, while the gains in uA are slightly smaller, they still present significant improvements in the geometric mean (with respect to all baselines) of 15.2\%, 1.4\%, and 2.1\% across the same models/datasets.
Additionally, the error rates across baselines exhibit small variance. Further, with EUAT,  we verify a larger separation of the uncertainty of the incorrect and correct predictions (Wasserstein distance), which highlights the robustness of our method across different tasks and domains, reaffirming its efficacy also in the challenging context of adversarial attacks.

%% file: sections/5.1_table.tex
\begin{table*}[t]
    \caption{Comparison of EUAT against the baselines using different evaluation metrics and considering four benchmarks.}
    \label{tab:results}
    \centering
    \begin{tabular}{llcccccc}
        \toprule
        \textbf{Benchmark} &\textbf{Baseline} & \textbf{uA} &  \textbf{uAUC} & \textbf{Corr. w/ res.} & \textbf{Wasser. dist.} & \textbf{ECE} & \textbf{Error} \\[3mm] \hline 
        \multirow{7}{*}{ResNet50/ImageNet}& \textbf{EUAT} & \textbf{0.804} & \textbf{0.878} & \textbf{0.655} & \textbf{0.301} & \textbf{0.223} &	\textbf{0.439} \\
         & \textbf{CE}           & 0.749 & 0.812 & 0.546 & 0.220 & 0.224 &	0.513 \\
         & \textbf{Calibration} & 0.743 & 0.807 & 0.538 & 0.193 & 0.274 &	0.535 \\
         & \textbf{DEUP}         & 0.580 & 0.591 & 0.291 & 0.030 & 0.429 &	0.522 \\
         & \textbf{Ensemble}     & 0.745 & 0.809 & 0.536 & 0.213 & 0.245 &	0.506 \\
         & \textbf{CALS}         & 0.746 & 0.809 & 0.542 & 0.214 & 0.235 &	0.525 \\
         & \textbf{CE+PE}        & 0.755 & 0.786 & 0.530 & 0.141 & 0.382 &	0.524 \\[1mm]
        \hline
        \multirow{7}{*}{Wide-ResNet/Cifar100}& \textbf{EUAT}  & \textbf{0.858} &	\textbf{0.891} &	\textbf{0.711}	& 0.216 & 0.162 &	0.273 \\
        & \textbf{CE}           & 0.794 &	0.774 &	0.546	& 0.128 & 0.235 &	0.296 \\
        & \textbf{Calibration} & 0.787 &	0.836 &	0.580	& \textbf{0.226} & 0.146 &	0.312 \\
        & \textbf{DEUP}         & 0.699 &	0.601 &	0.300	& 0.052 & 0.252 &	0.328 \\
        & \textbf{Ensemble}     & 0.742 &	0.696 &	0.466	& 0.089 & 0.294 &	0.332 \\
        & \textbf{CALS}         & 0.813 &	0.831 &	0.589	& 0.233 & \textbf{0.114} &	\textbf{0.264} \\
        & \textbf{CE+PE}        & 0.779 &	0.740 &	0.518	& 0.099 & 0.252 &	0.300 \\[1mm]
        \hline
        \multirow{7}{*}{ResNet18/Cifar10}& \textbf{EUAT}& 0.914 &	\textbf{0.921} & \textbf{0.626} &	\textbf{0.410} &	0.018 &	0.099 \\
        & \textbf{CE}          & 0.905 &	0.866 & 0.576 &	0.273 &	0.025 &	0.103 \\
        & \textbf{Calibration}& 0.898 &	0.893 & 0.533 &	0.329 &	0.030 &	0.108 \\
        & \textbf{DEUP}        & 0.917 &	0.563 & 0.297 &	0.032 &	0.039 &	0.094 \\
        & \textbf{Ensemble}    & \textbf{0.919} &	0.840 & 0.545 &	0.224 &	0.039 &	\textbf{0.087} \\
        & \textbf{CALS}        & 0.907 &	0.875 & 0.563 &	0.283 &	\textbf{0.012} &	0.101 \\
        & \textbf{CE+PE}       & 0.907 &	0.837 & 0.566 &	0.213 &	0.052 &	0.102 \\[1mm]
        \hline 
        \multirow{7}{*}{ResNet18/SVHN}& \textbf{EUAT}  & 0.960 &	\textbf{0.927}	& \textbf{0.638} &	\textbf{0.479} & \textbf{0.011} & 0.047 \\
        & \textbf{CE}           & 0.956 &	0.841	& 0.572 &	0.232 & 0.021 & 0.047 \\
        & \textbf{Calibration} & 0.953 &	0.902	& 0.537 &	0.346 & 0.026 & 0.051 \\
        & \textbf{DEUP}         & 0.960 &	0.564	& 0.312 &	0.040 & 0.024 & 0.044 \\
        & \textbf{Ensemble}     & 0.960 &	0.756	& 0.515 &	0.162 & 0.029 & 0.043 \\
        & \textbf{CALS}         & \textbf{0.963} &	0.867	& 0.569 &	0.264 & \textbf{0.011} & \textbf{0.041} \\
        & \textbf{CE+PE}        & 0.959 &	0.799	& 0.547 &	0.184 & 0.029 & 0.045 \\[1mm]
        \bottomrule
    \end{tabular}
    \vspace{2mm}
    
\end{table*}

%% file: sections/5.1_plots.tex
\begin{figure*}[t]
\centering
\captionsetup{justification=centering}
    \begin{subfigure}[h]{0.45\textwidth}
        \includegraphics[clip, trim=520mm 4mm 0mm 205mm,  height=0.5\textwidth]{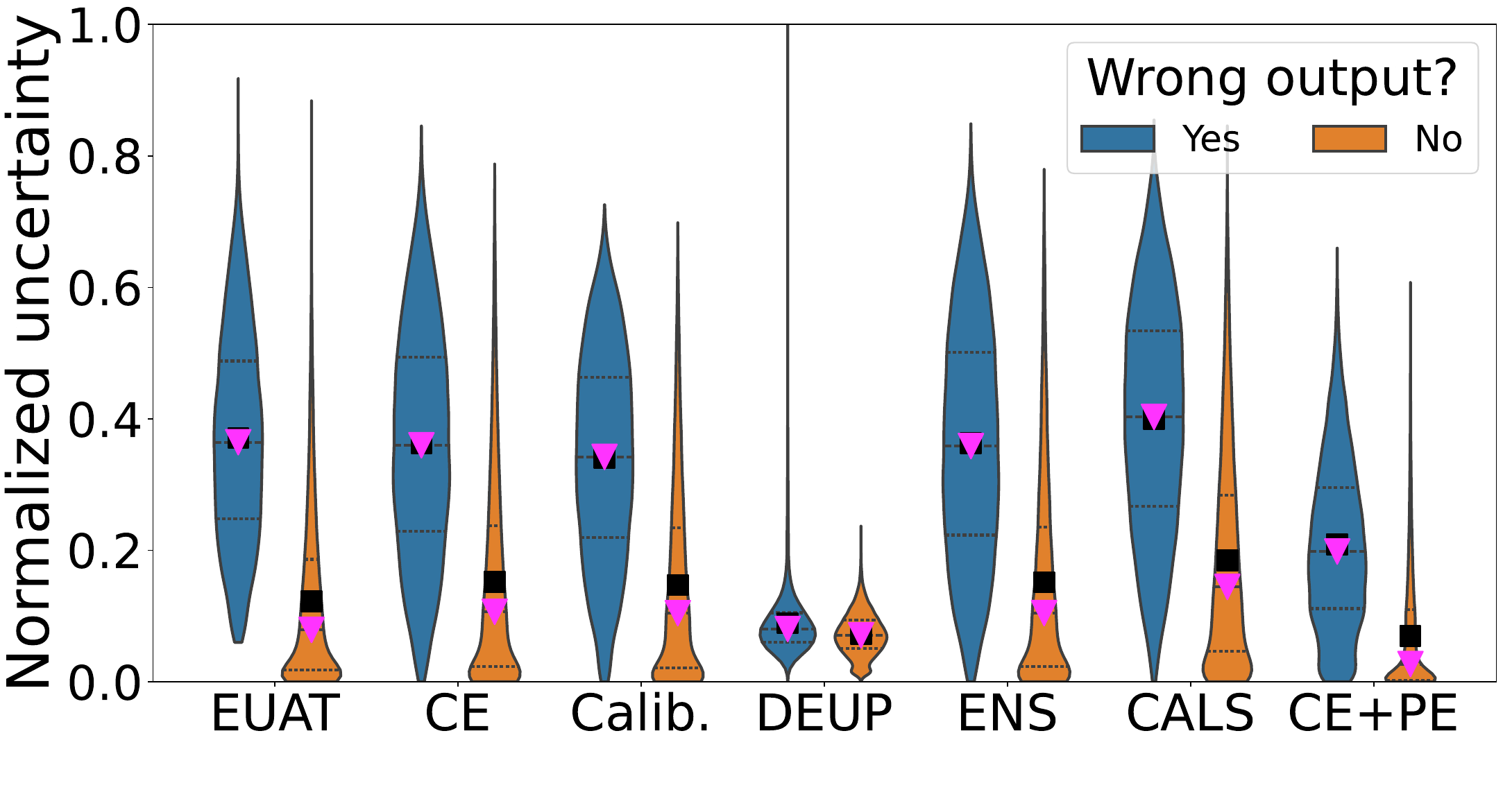}
        \caption{ResNet50/ImageNet}
        \label{fig:unc2_imgNet_std}
    \end{subfigure}
    \begin{subfigure}[h]{0.45\textwidth}
        \includegraphics[clip, trim=510mm 4mm 14mm 205mm,  height=0.5\textwidth]{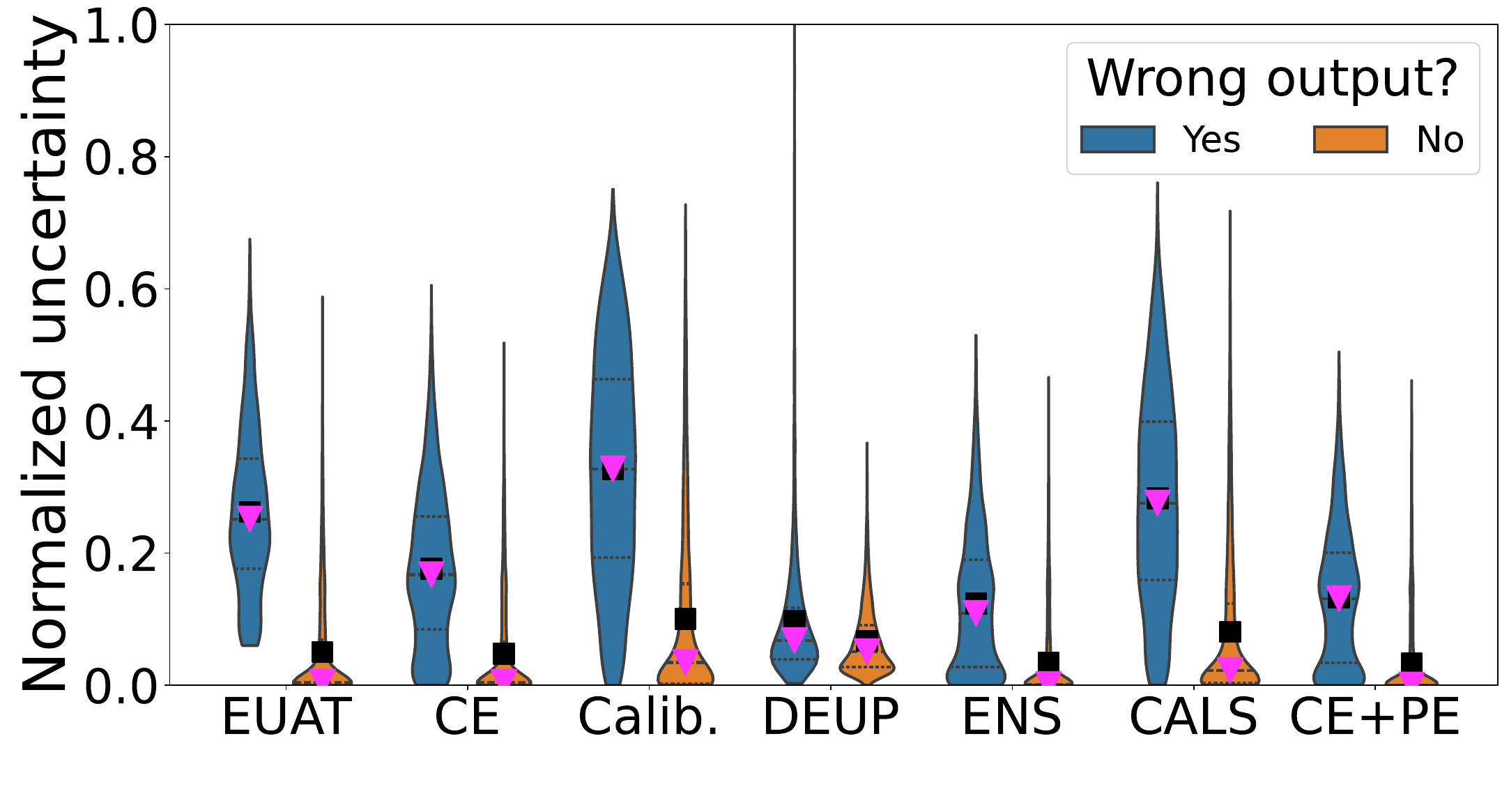}
        \caption{Wide-ResNet/Cifar100}
        \label{fig:unc2_cifar100_std}
    \end{subfigure}
\vspace{3mm}

    \begin{subfigure}[h]{0.45\textwidth}
        \includegraphics[clip, trim=520mm 4mm 14mm 205mm, height=0.5\textwidth]{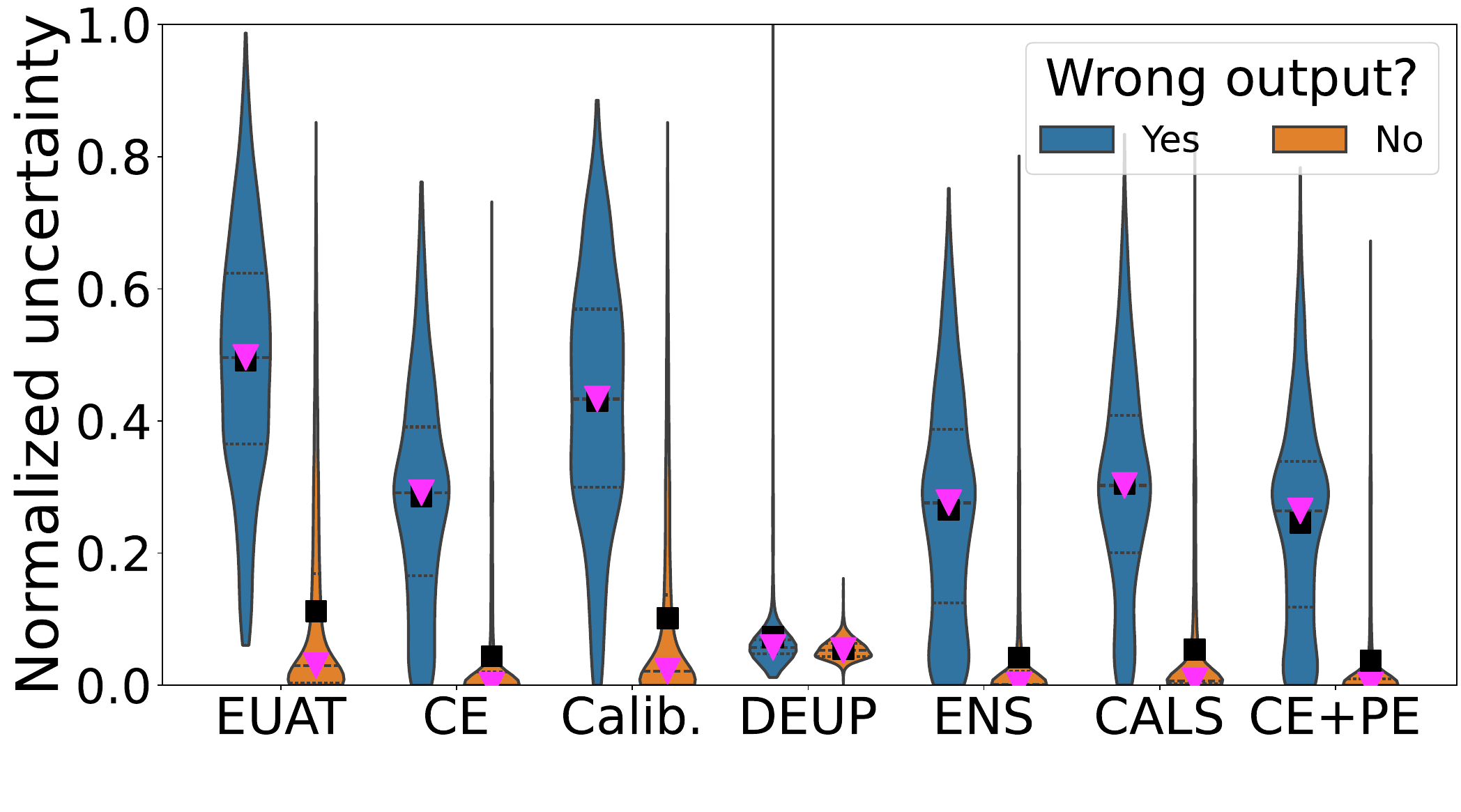}
        \caption{ResNet18/Cifar10}
        \label{fig:unc2_cifar10_std}
    \end{subfigure}
    \begin{subfigure}[h]{0.45\textwidth}
        \includegraphics[clip, trim=520mm 4mm 2mm 205mm,  height=0.5\textwidth]{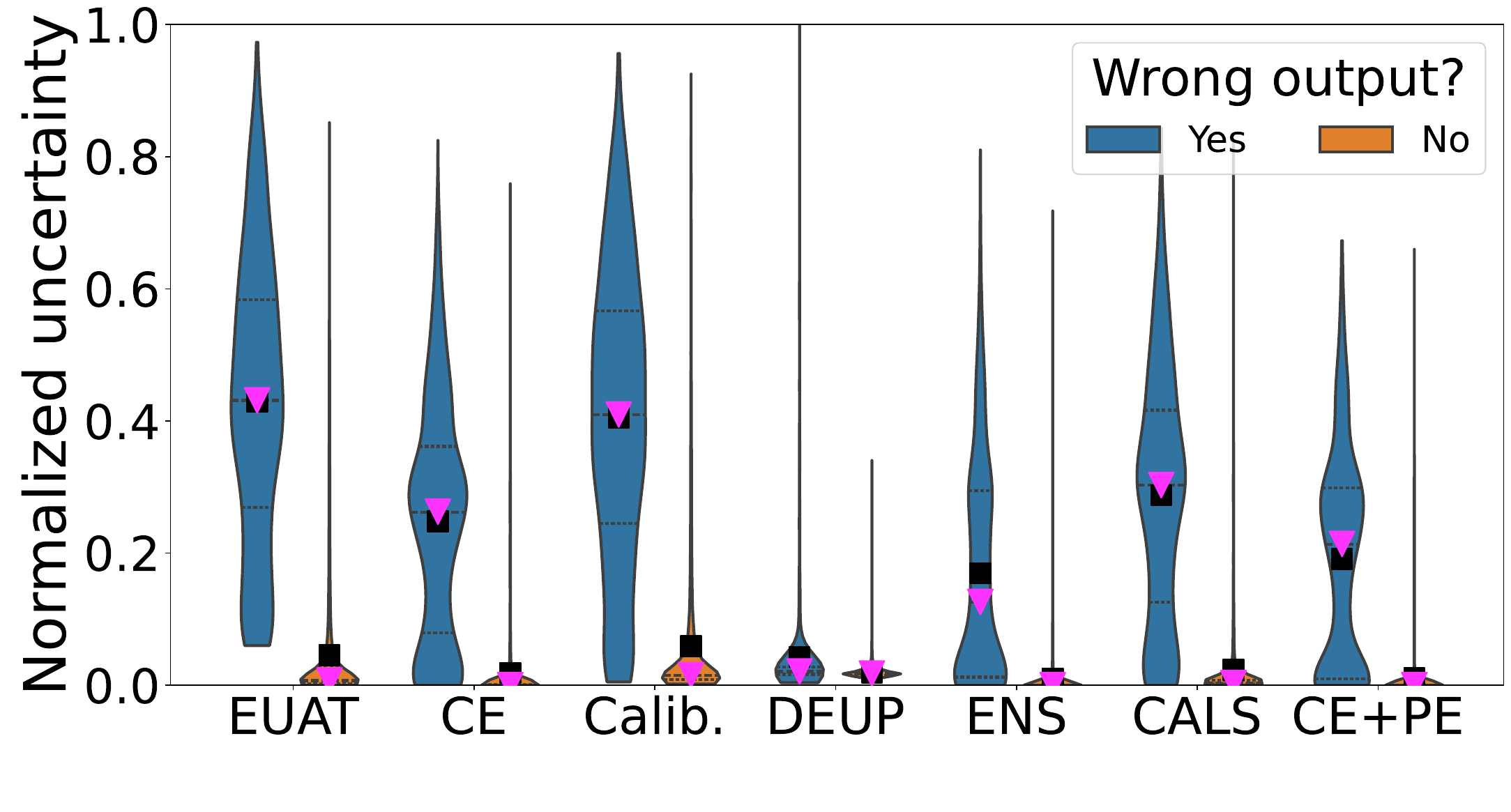}
        \caption{ResNet18/SVHN}
        \label{fig:unc2_svhn_std}
    \end{subfigure}
\vspace{4mm}
    \captionsetup{justification=justified}
    \caption{Normalized uncertainty distribution of correct and incorrect predictions for the different baselines (the average value of each distribution is marked with a black square, and the median with a pink triangle).}
     \label{fig:results}
\vspace{4mm}
\end{figure*}

%% file: sections/5.2_binary.tex
\begin{table*}[t]
    \caption{Comparison of EUAT against the baselines using different evaluation metrics and considering a binary classification problem.}
    \label{tab:binaryCifar10}
    \centering
    \begin{tabular}{lccccccccccc}
        \toprule
        \multirow{2}{*}{\textbf{Baseline}} & \multirow{2}{*}{\textbf{uA}} &  \multirow{2}{*}{\textbf{uAUC}} & \textbf{Corr.} & \textbf{Wasser.} & \multirow{2}{*}{\textbf{ECE}} & \textbf{Error} & \textbf{Error}	& \multirow{2}{*}{\textbf{F1}} & \multirow{2}{*}{\textbf{Precision}} & \multirow{2}{*}{\textbf{TPR}} & \multirow{2}{*}{\textbf{TNR}} \\
        &  & & \textbf{w/ res. } & \textbf{dist.} & & \textbf{w/o flip} & \textbf{w/ flip} & & & & \\
        [1mm] \hline
    \textbf{EUAT} & \textbf{0.861} & \textbf{0.816} & \textbf{0.446} & \textbf{0.405} & \textbf{0.112} & 0.152 & \textbf{0.139} & \textbf{0.860} & \textbf{0.863} & \textbf{0.858} & 0.864 \\
    \textbf{CE}   & 0.845 & 0.759 & 0.435 & 0.350 & 0.113 & 0.152 & 0.221 & 0.763 & 0.823 & 0.712 & 0.847\\
    \textbf{Calib.} & 0.783 & 0.779 & 0.366 & 0.356 & 0.313 & 0.159 & 0.290 & 0.733 & 0.670 & 0.809 & 0.603 \\
    \textbf{DEUP} & 0.834 & 0.534 & 0.209 & 0.023 & 0.124 & 0.172 & 0.178 & 0.812 & 0.859 & 0.771 & \textbf{0.874} \\
    \textbf{Ensemble} & \textbf{0.861} & 0.683 & 0.358 & 0.216 & \textbf{0.112} & \textbf{0.147} & 0.163 & 0.834 & 0.847 & 0.822 & 0.852 \\
    \textbf{CALS} & 0.831 & 0.748 & 0.407 & 0.326 & 0.119 & 0.164 & 0.163 & 0.834 & 0.847 & 0.822 & 0.852\\
    \textbf{CE+PE}& 0.811 & 0.690 & 0.360 & 0.274 & 0.153 & 0.184 & 0.225 & 0.759 & 0.817 & 0.709 & 0.842\\[1mm]
        \bottomrule
    \end{tabular}
    \vspace{4mm}
    
\end{table*}

%% file: sections/5.3_ood.tex
\begin{table}[t]
    \caption{Comparison of EUAT against the baselines using different evaluation metrics and tested with out-of-distribution samples with Cifar10.}
    \label{tab:ood}
    \centering
    \addtolength{\tabcolsep}{-0.1em}
    \begin{tabular}{lcccccc}
        \toprule
        \multirow{2}{*}{\textbf{Baseline}} & \multirow{2}{*}{\textbf{uA}} &  \multirow{2}{*}{\textbf{uAUC}} & \textbf{Corr.} & \textbf{Wasser.} & \multirow{2}{*}{\textbf{ECE}} & \multirow{2}{*}{\textbf{Error}} \\
        &  & & \textbf{w/ res. } & \textbf{dist.} & & \\
        [1mm] \hline
        \textbf{EUAT}  & \textbf{0.754} & \textbf{0.796} & \textbf{0.529} & \textbf{0.255} & \textbf{0.143} & 0.489\\
        \textbf{CE}   & 0.691 & 0.676 & 0.311 & 0.126 & 0.292 & 0.539\\
        \textbf{Calib.}& 0.619 & 0.553 & 0.110 & 0.036 & 0.497 & 0.619\\
        \textbf{DEUP}   & 0.617 & 0.509 & 0.044 & 0.003 & 0.551 & 0.617\\
        \textbf{Ensemble} & 0.753 & 0.734 & 0.426 & 0.177 & 0.237 & 0.555\\
        \textbf{CALS}  & 0.686 & 0.716 & 0.383 & 0.147 & 0.216 & \textbf{0.464}\\
        \textbf{CE+PE} & 0.681 & 0.663 & 0.299 & 0.103 & 0.418 & 0.568\\[1mm]
        \bottomrule
    \end{tabular}
    \vspace{4mm}
\end{table}

%% file: sections/5.4_tableAT.tex
\begin{table*}[t]
    \caption{Comparison of EUAT against the baselines considered using different evaluation metrics in the adversarial training scenario 
    using  three benchmarks.}
    \label{tab:results_at}
    \centering
    \addtolength{\tabcolsep}{-0.05em}
    \begin{tabular}{ll|cccccc|cccccc}
        \toprule
          & &\multicolumn{6}{c|}{\textbf{Evaluation with clean data}}&\multicolumn{6}{c}{\textbf{Evaluation with adversarial data}}\\ [2mm] 
         \multirow{2}{*}{\textbf{Benchmark}} &\multirow{2}{*}{\textbf{Baseline}} & \multirow{2}{*}{\textbf{uA}} &  \multirow{2}{*}{\textbf{uAUC}} & \textbf{Corr. w/} & \textbf{Wasser.} & \multirow{2}{*}{\textbf{ECE}} & \multirow{2}{*}{\textbf{Error}} & \multirow{2}{*}{\textbf{uA}} &  \multirow{2}{*}{\textbf{uAUC}} & \textbf{Corr. w/} & \textbf{Wasser.} & \multirow{2}{*}{\textbf{ECE}} & \multirow{2}{*}{\textbf{Error}} \\
        &&&& \textbf{res. } & \textbf{dist.} &&&&& \textbf{res } & \textbf{dist.} & & \\ [2mm] \hline 
        %
        \multirow{7}{*}{Cifar100}& 
        \textbf{EUAT} & \textbf{0.850} & \textbf{0.902} & \textbf{0.709} & 0.254 & 0.154 & 0.303 & 0.791 & \textbf{0.853} & 0.643 & \textbf{0.224} & 0.268 & \textbf{0.436}\\
        & \textbf{CE} & 0.786 & 0.793 & 0.565 & 0.154 & 0.241 & 0.329 & 0.694 & 0.716 & 0.430 & 0.104 & 0.377 & 0.460 \\
        & \textbf{Calibration} & 0.771 & 0.839 & 0.576 & \textbf{0.271} & \textbf{0.087} & 0.335 & 0.734 & 0.801 & 0.529 & 0.222 &\textbf{ 0.202} & 0.456\\
        & \textbf{DEUP} & 0.677 & 0.585 & 0.222 & 0.036 & 0.235 & 0.337 & 0.577 & 0.584 & 0.214 & 0.033 & 0.391 & 0.493\\
        & \textbf{Ensemble} & 0.800 & 0.817 & 0.597 & 0.161 & 0.183 & \textbf{0.288} & \textbf{0.865} & \textbf{0.852} & \textbf{0.750} & 0.194 & 0.334 & \textbf{0.436}\\
        & \textbf{CALS} & 0.788 & 0.787 & 0.567 & 0.159 & 0.245 & 0.332 & 0.689 & 0.714 & 0.430 & 0.105 & 0.383 & 0.466\\
        & \textbf{CE+PE}& 0.773 & 0.750 & 0.535 & 0.113 & 0.286 & 0.354 & 0.625 & 0.633 & 0.346 & 0.059 & 0.422 & 0.482\\[1mm]
        \hline
        \multirow{7}{*}{Cifar10}& 
        \textbf{EUAT}          & 0.905 & \textbf{0.921} & \textbf{0.635} & \textbf{0.414} & 0.031 & 0.117 & 0.845 & \textbf{0.883}  & 0.601 & \textbf{0.353} & 0.031	& 0.196 \\
        & \textbf{CE}          & 0.895 & 0.898 & 0.571 & 0.306 & 0.011 & 0.116 & 0.826 & 0.847  & 0.542 & 0.259 & 0.028	& 0.198\\
        & \textbf{Calibration} & 0.890 & 0.893 & 0.528 & 0.339 & 0.051 & 0.119 & 0.819 & 0.845  & 0.513 & 0.278 & 0.019	& 0.201\\
        & \textbf{DEUP}        & 0.904 & 0.591 & 0.235 & 0.044 & 0.018 & 0.104 & 0.804 & 0.603	& 0.237 & 0.034 & 0.092	& 0.205\\
        & \textbf{Ensemble}    & \textbf{0.923} & 0.894 & 0.535 & 0.310 & \textbf{0.007} & \textbf{0.081} & \textbf{0.901} & 0.881  & \textbf{0.780} & 0.348 & 0.083	& \textbf{0.179}\\
        & \textbf{CALS}        & 0.895 & 0.898 & 0.579 & 0.317 & 0.009 & 0.117 & 0.833 & 0.842  & 0.542 & 0.264 & \textbf{0.015} & 0.196\\
        & \textbf{CE+PE}       & 0.895 & 0.850 & 0.564 & 0.232 & 0.046 & 0.119 & 0.819 & 0.789  & 0.505 & 0.177 & 0.104 & 0.197\\[1mm]
        \hline 
       \multirow{7}{*}{SVHN}& 
        \textbf{EUAT}          & 0.944 & \textbf{0.947} & \textbf{0.649} & \textbf{0.447} & 0.030 & 0.072 & 0.794 & \textbf{0.848} & 0.588 & \textbf{0.307} & \textbf{0.020} & 0.281 \\
        & \textbf{CE}          & 0.931 & 0.893 & 0.590 & 0.335 & \textbf{0.009} & 0.080 & 0.749 & 0.773 & 0.473 & 0.196 & 0.134 & 0.307 \\
        & \textbf{Calibration} & 0.929 & 0.916 & 0.521 & 0.428 & 0.138 & 0.082 & 0.758 & 0.815 & 0.516 & 0.261 & \textbf{0.020} & 0.299 \\
        & \textbf{DEUP}        & 0.933 & 0.560 & 0.258 & 0.035 & 0.029 & 0.071 & 0.667 & 0.535 & 0.302 & 0.022 & 0.313 & 0.354 \\
        & \textbf{Ensemble}    & \textbf{0.951} & 0.893 & 0.545 & 0.333 & \textbf{0.009} & \textbf{0.053} & \textbf{0.881} & 0.827 & \textbf{0.681} & 0.221 & 0.234 & 0.308 \\
        & \textbf{CALS}        & 0.940 & 0.894 & 0.592 & 0.340 & \textbf{0.009} & 0.069 & 0.820 & 0.757 & 0.481 & 0.206 & 0.117 & \textbf{0.200} \\
        & \textbf{CE+PE}       & 0.932 & 0.874 & 0.596 & 0.274 & 0.021 & 0.078 & 0.823 & 0.736 & 0.449 & 0.145 & 0.127 & 0.187 \\[1mm]
        \bottomrule
    \end{tabular}
    \vspace{4mm}
    
\end{table*}

%% file: sections/6_conclusion.tex
\section{Conclusion and Future Work}
\label{sec:conclusion}

This paper introduces Error-Driven Uncertainty Aware Training, a novel approach designed to refine the estimation of model uncertainty. EUAT is engineered to achieve two primary objectives: first, to heighten uncertainty when models generate inaccurate predictions, and second, to output low uncertainty when predictions are correct.
This dual-purpose strategy is achieved through the usage of two loss functions, which vary depending on whether training examples are correctly or incorrectly predicted by the model. By minimizing uncertainty for accurate predictions and maximizing it for mispredictions, while striving to minimize error rates, EUAT aims to enhance model's trustworthiness.


We evaluate EUAT using six different baselines and considering six metrics. Our evaluation encompasses diverse problems, including image recognition, binary classification, out-of-distribution detection, and adversarial training settings. Across all the evaluated domains, EUAT demonstrates an enhanced ability to differentiate between erroneous and accurate predictions based on uncertainty levels (evaluated via the Wasserstein distance), thereby increasing model trustworthiness.
More in detail, the experimental results showcase EUAT's superior performance across the majority of the considered cases.
Further, even in settings where EUAT is not the most competitive solution, it still achieves comparable performance to the best performing baselines.



The EUAT method has been designed for classification tasks, and evaluated using image recognition models. In the future, we intend to broaden the applicability of EUAT to additional domains such as regression problems or natural language processing tasks. 